\documentclass[final]{svjour3PP}

\usepackage{amsfonts,amsmath,amssymb}

\usepackage{mathptmx}
\usepackage[scr=cm]{mathalpha}
\usepackage{cite}

\AtBeginDocument{%
  \setlength{\oddsidemargin}{\dimexpr(\paperwidth-\textwidth)/2-1in}%
  \setlength{\evensidemargin}{\oddsidemargin}%
  \setlength{\topmargin}{%
  \dimexpr(\paperheight-\textheight)/2-\headheight-\headsep-1in}%
}

\def\be{\begin{equation}}
	\def\bea{\begin{eqnarray*}}
		\def\ee{\end{equation}}
	\def\eea{\end{eqnarray*}}
\def\ba{\begin{array}}
	\def\ea{\end{array}}
\def\bi{\begin{itemize}}
	\def\ei{\end{itemize}}

\def\bfR{\mathbf{R}}

\usepackage{graphicx}
\usepackage{a4,amssymb,amsmath}

\title{Convergence analysis of online algorithms for vector-valued kernel regression}
\author{Michael Griebel \and Peter Oswald}
\institute{M. Griebel\at Institute for Numerical Simulation, Universit\"at Bonn, Friedrich-Hirzebruch-Allee 7, 53115 Bonn, Germany, and Fraunhofer Institute for Algorithms and Scientific Computing (SCAI), Schloss Birlinghoven, 53754 Sankt Augustin, Germany\\
Corresponding author, tel.: +49-228-7369829, \email{griebel@ins.uni-bonn.de}
\and P. Oswald \at Institute for Numerical Simulation, Universit\"at Bonn, Friedrich-Hirzebruch-Allee 7, 53115 Bonn, Germany\\
\email{agp.oswald@gmail.com}}
\titlerunning{Online algorithms and vector-valued kernel regression}
\authorrunning{M. Griebel \and P. Oswald}

\date{}
\begin{document}
\maketitle
\begin{abstract}
We consider the problem of approximating the regression function $f_\mu:\, \Omega \to Y$ from noisy $\mu$-distributed  vector-valued data $(\omega_m,y_m)\in\Omega\times Y$ by an online learning algorithm using a reproducing kernel Hilbert space $H$ (RKHS) as prior. In an online algorithm, i.i.d. samples become available one by one via a random process and are successively processed to build approximations to the regression function. 
Assuming that the regression function essentially belongs to $H$ (soft learning scenario), we provide estimates for the expected squared error in the RKHS norm of the approximations $f^{(m)}\in H$ obtained by a standard regularized online approximation algorithm. In particular, we show an order-optimal estimate 
$$
\mathbb{E}(\|\epsilon^{(m)}\|_H^2)\le C (m+1)^{-s/(2+s)},\qquad m=1,2,\ldots,
$$
where $\epsilon^{(m)}$ denotes the error term after $m$ processed data, the parameter $0<s\leq 1$ expresses an additional smoothness assumption on the regression function, and the constant $C$ depends on the variance of the input noise, the smoothness of the regression function, and other parameters of the algorithm.
The proof, which is inspired by results on Schwarz iterative methods \cite{GrOs2018} in the noiseless case, uses only elementary Hilbert space techniques and minimal assumptions on the noise, the feature map that defines $H$ and the associated covariance operator.
\end{abstract}

\section{Introduction}\label{sec1}
In this paper, we consider the problem of learning the regression function from noisy vector-valued data using an appropriate RKHS as a prior. For relevant background on the theory of kernel methods, see \cite{Va1998,ScSm2002,CuSm2002,CuZh2007,StCh2008} and especially \cite{MiPo2005,CVT2006,CVTU2010} in the vector-valued case. Our focus is to obtain estimates for the expectation of the squared error norm in the RKHS $H$ of approximations to the regression function. These approximations are constructed in an incremental way by so-called online algorithms. The setting we use is as follows: Let be given $N\le \infty$ samples
$(\omega_m,y_m) \in \Omega\times Y$, $m=0,\ldots,N-1$, of an input-output process $\omega\to y$, which are i.i.d. with respect to a (generally unknown) probability measure $\mu$ defined on $\Omega\times Y$. Let $\Omega$ be a compact metric space, $Y$ be a separable Hilbert space, and $\mu$ be a Borel measure. We are looking for a regression function $f_\mu:\,\Omega \to Y$ that optimally represents the underlying input-output process in some sense. Algorithms for least-squares regression aim to find approximations to the solution
$$
f_\mu(\omega) =\mathbb{E}(y|\omega)\in L^2_\rho(\Omega,Y)
$$
of the minimization problem
\be\label{Min}
\mathbb{E}(\|f(\omega) - y\|_Y^2)=\int_{\Omega\times Y} \|f(\omega) - y\|_Y^2\,d\mu(\omega,y) \longmapsto \min
\ee
for $f\in L^2_\rho(\Omega,Y)$ from
the samples $(\omega_m,y_m), m=0,\ldots,N-1$, where $\rho(\omega)$ 
is the marginal probability measure generated by $\mu(\omega,y)$ on $\Omega$.\footnote{The symbol $\mathbb{E}$ denotes expectations of random variables with respect to the underlying probability space, which may vary from formula to formula but should be clear from the context.}
For the minimization problem (\ref{Min}) to be meaningful, one needs
$$
\mathbb{E}(\|y\|_Y^2)=\int_{\Omega\times Y} \|y\|_Y^2\,d\mu(\omega,y) =\int_\Omega \mathbb{E}(\|y\|_Y^2|\omega) \,d\rho(\omega) < \infty.
$$
Since solving the discretized least-squares problem
\be\label{Min1}
\frac1N \sum_{m=0}^{N-1} \|f(\omega_m) - y_m\|^2_Y \longmapsto \min
\ee
for $f\in L^2_\rho(\Omega,Y)$ is an ill-posed problem that makes no sense without further regularization,
it is customary to add a prior assumption $f\in H$, where $H\subset L^2_\rho(\Omega,Y)$ is a set of functions $f:\;\Omega\to Y$ such that point evaluations $\omega\to f(\omega)$ are continuous maps. Staying with the Hilbert space setting, candidates for $H$ are  vector-valued RKHS 
$$ 
H=\{f_v(\omega)=R^\ast_\omega v:\;\|f_v\|_H:=\|v\|_V,\;v\in V\}
$$ 
introduced by means of a family $\mathbf{R}:=\{R_\omega\}_{\omega\in\Omega}$ of bounded linear operators $R_\omega:\,Y\to V$ mapping into a separable Hilbert space $V$ called feature space.
Detailed definitions and necessary assumptions on the feature map $\mathbf{R}$ are given in the next section.\footnote{In the literature \cite{MiPo2005},\cite[Chapter 4]{StCh2008}, feature maps are typically introduced as operator families acting from $V$ to $Y$, which, in our case, would be the family $\mathbf{R}^\ast$ of Hilbert adjoints $R_\omega^\ast:\,V\to Y$. For the purpose of this paper, we stay close to the notation used in \cite{GrOs2018} and call $\mathbf{R}$ feature map, see also \cite{CVT2006}.} Obviously, if $f_v(\omega)=0$ for all $\omega\in\Omega$ implies $v=0$ then $\mathbf{R}^\ast$ is an isometry and the RKHS $H$ and the feature space $V$  can be identified. 

The online algorithms considered in this paper start from an initial guess $u^{(0)}\in V$ and build a sequence of 
successive approximations $u^{(m)}\in V$, where $u^{(m+1)}$ is a linear combination of the previous
approximation $u^{(m)}$ and a term that includes the residual $y_m-R^\ast_{\omega_m}u^{(m)}$ with respect to the
currently processed sample $(\omega_m,y_m)$. More precisely, the
update formula in the feature space $V$ can be written in the form
\be\label{OnlineV}
u^{(m+1)}(\omega)=\alpha_m(u^{(m)}+\mu_m R_{\omega_m}(y_m-R^\ast_{\omega_m}u^{(m)})), \qquad m=0,1,\ldots, N-1.
\ee
By setting $f^{(m)}:=f_{u^{(m)}}$, the corresponding iteration in the RKHS $H$ 
has the equivalent form
\be\label{OnlineH}
f^{(m+1)}(\omega)=\alpha_m(f^{(m)}(\omega)+\mu_m K(\omega,\omega_m)(y_m-f^{(m)}(\omega_m))), \qquad m=0,1,\ldots, N-1,
\ee
where $K(\omega,\theta):\,Y\to Y$, $\omega,\theta\in \Omega$, is the operator kernel of the RKHS determined by the feature map.
\footnote{The formula (\ref{OnlineH}) shows that, in order to execute the algorithm, the explicit use of $V$ and the feature map $\mathbf{R}$ can be avoided if the operator kernel $K$ and not the feature map is given, which is often the case. In the convergence proofs, it is more convenient to use (\ref{OnlineV}).} 
The parameters $\alpha_m, \mu_m$ are specifically given by
\be\label{OnlineA}
\alpha_m=\frac{m+1}{m+2},\qquad  \mu_m=\frac{A}{(m+1)^t},\qquad m=0,1,\ldots, 
\ee
where the constants $1/2<t<1$ and $A>0$ will be properly fixed later. Following tradition,
$\alpha_m$ and $\mu_m$ are called regularization and step-size parameters, respectively. 

Our main result, namely Theorem~\ref{theo1}, concerns a sharp asymptotic estimate for the expected squared error 
$$
\mathbb{E}(\|u-u^{(m)}\|_V^2) =\mathbb{E}(\|f_u-f^{(m)}\|_H^2), \qquad m\to \infty,
$$
of these iterations in the $V$ and $H$ norms, respectively. Here
$u\in V$ is the unique minimizer of the minimization problem\footnote{The existence of such a $u$ is a strong assumption necessary for investigating convergence in the RKHS norm	and is called \emph{soft learning scenario} in the literature, see, e.g., \cite{FiSt2020}. Our method of proof does not automatically extend to the \emph{hard learning scenario}, where only $f_\mu\in L^2_\rho(\Omega,Y)$ is assumed and $L^2_\rho(\Omega,Y)$ convergence is studied.}
\be\label{Min2}
J(v):= \mathbb{E}(\|f_v-y\|_Y^2)=\int_{\Omega\times Y} \|R^\ast_\omega v-y\|_Y^2\,d\mu(\omega,y) \longmapsto \min.
\ee
These estimates hold under standard assumptions on the feature map $\mathbf{R}$ and the operator kernel $K$, the
parameters in (\ref{OnlineA}) and the smoothness~$s$ of $u\in V$ measured in a scale of smoothness spaces $V_{P_\rho}^s\subset V$ associated with the underlying covariance operator $P_\rho=\mathbb{E}(R_\omega R^\ast_\omega)$. 

Our approach is an extension of earlier work \cite{GrOs2018} on Schwarz iterative methods in the noiseless case, where $y_m=f_u(\omega_m)=R^\ast_{\omega_m} u$ was assumed. In our opinion, the main contribution of this paper is providing a proof of a generic
convergence result for online learning in the general vector-valued case which uses only basic Hilbert space techniques. In particular, compactness assumptions on the covariance operator  $P_\rho$ do not play a role other than to simplify the presentation of proofs, and structural assumptions on the noise $y-f_\mu(\omega)$ are not needed. We note that the online learning version studied in this paper can be reinterpreted as the iterative solution of an inverse problem for the equation $R_\omega^\ast u = y$ (equality in $L^2_\rho(\Omega,Y)$) with random right-hand side~$y$. Iterative regularization of inverse problems with deterministic noise is studied in \cite[Chapter 6]{EHN2000} using similar tools. The connection between kernel regression and statistical inverse problems is well-known, see \cite{VRCGO2005}, but we are not aware of a result similar to our Theorem \ref{theo1} covering the online case. More sophisticated tools (for instance, concentration inequalities valid under certain assumptions on the noise, integral operator techniques, taking into account decay properties of the spectrum of $P_\rho$) lead to stronger results that apply to the hard learning scenario as well. Furthermore, we do not claim relevance for particular applications. Motivation for the vector-valued case comes from multitask learning (here, $Y=\mathbb{R}^d$) and functional learning, see the references in \cite{ARL2012,KDPCRA2015,LMMG2024}. In most of the current applications, constructions involving $\mathbb{R}$-valued kernels are used to cover the vector-valued case. In Section \ref{sec2}, we provide an example with $Y=H^1_0(G)$ following \cite{KDPCRA2015}.

The remainder of this paper is organized as follows:
In Section \ref{sec11} we introduce vector-valued RKHS in terms of feature maps $\mathbf{R}$ and spaces $V$, define $P_\rho$ and the associated scale of smoothness spaces $V_{P_\rho}^s$, and discuss properties of the minimization problem (\ref{Min2}). 
This sets the stage for the analysis of our online learning algorithm in $V$ and allows us to formulate our main convergence result, namely Theorem~\ref{theo1}. In Section \ref{sec2} we review related results from the literature. In Section \ref{sec3} we then provide the detailed proof of Theorem \ref{theo1}. In Section \ref{sec4} we give further remarks on Theorem \ref{theo1}, show the 
divergence in expectation 
$$
\|\mathbb{E}(u^{(m)})\|_V \to \infty, \qquad m\to \infty,
$$
of the online algorithm (\ref{OnlineV})-(\ref{OnlineA}) if (\ref{Min2}) does not
possess a unique minimizer $u\in V$, and consider a simple special case of "learning" an element $u$ of a Hilbert space $V$ from noisy measurements of its coefficients with respect to a complete orthonormal system (CONS) in $V$. 

\section{Setting and main result}\label{sec11}
Let us first introduce our approach to vector-valued RKHS $H\subset L^2_\rho(\Omega,Y)$, where $\Omega$ is a compact metric space with Borel probability measure $\rho$ and $Y$ is a separable Hilbert space.\footnote{The facts stated below without proof can be found in \cite{CVT2006,PaRa2016}. The assumptions on $\Omega$ can be weakened.} In the learning problem under consideration, $\rho$ is the marginal measure induced from a Borel probability measure $\mu$ on $\Omega\times Y$. Such an RKHS $H$ can be implicitly introduced by a family $\bfR=\{R_\omega\}_{\omega\in \Omega}$ of bounded linear operators $R_\omega:\,Y\to V$, called feature map, where $V$ is another separable Hilbert space (we will tacitly assume that $V$ is infinite-dimensional). More precisely, we introduce the notation
$$
f_v(\omega):=R_\omega^\ast v, \quad \omega\in \Omega,\quad v\in V,
$$ 
and set 
$$
H:=\{f_v:\;v\in V,\quad\|f_v\|_H:=\|v\|_V\}.
$$
Without loss of generality, we assume that $f_v(\omega)=0$ for all $\omega\in \Omega$ implies $v=0$
(otherwise, replace $V$ by its subspace $(\cap_{\omega\in \Omega}\mathrm{ker}(R_\omega^\ast))^\perp$).
That is, the RKHS $H$ and the space $V$ can be identified, which allows us to easily switch between $H$ and $V$ in the sequel.

We impose the following conditions on the feature map $\bfR$: We assume \emph{uniform boundedness}
\be\label{AssR2} 
\|R_{\omega}\|_{Y\to V}^2\le \Lambda <\infty,\qquad \omega\in \Omega,
\ee  
with some $\Lambda<\infty$ and \emph{(weak) measurability}, i.e., the scalar function
$(R_\omega y,v)_V$ is Borel measurable on $\Omega$ for each pair $(y,v)\in Y\times V$.
By (\ref{AssR2}), the functions $f_v\in H$ are bounded on $\Omega$ which, together with the
measurability assumption, implies $H\subset L^2(\Omega,Y)$ for any Borel probability measure $\rho$ on $\Omega$.\footnote{For simplicity, we silently identify $f_v$ with its equivalence class, whenever this is formally necessary.}

The condition (\ref{AssR2}) is equivalent to the uniform boundedness
\be\label{AssR2a}
\|K(\omega,\theta)\|_{Y}\le \Lambda,\qquad \omega,\theta \in\Omega,
\ee
of the operator kernel 
$$
K(\omega,\theta):=R_\omega^\ast R_\theta:\;Y\to Y,\qquad \omega,\theta\in \Omega,
$$
associated with the vector-valued RKHS $H$. Furthermore, (\ref{AssR2}) is equivalent to the uniform boundedness
of the operator family 
$$
P_\omega:=R_\omega R_\omega^\ast:\, V\to V,\qquad \omega\in \Omega,
$$ 
in $V$,
i.e.,
$$
\|P_\omega\|_V\le \Lambda,\qquad \omega\in\Omega.
$$ 
Moreover, weak measurability of the operator families $K(\omega,\theta)$, $R^\ast_\omega$ and $P_\omega$ on $\Omega\times\Omega$ and $\Omega$, respectively, follows then from the weak measurability of $\mathbf{R}$ as well.
This ensures that Bochner integrals of functions from $\Omega$ into $Y$ and $V$, respectively, which appear in the formulas below, are well-defined. 

For fixed $V$ and $\bfR$ satisfying the above properties, instead of solving the minimization problem (\ref{Min}) on $L^2_\rho(\Omega,Y)$, 
one now searches for the minimizer $u\in V$ of (\ref{Min2}).
The solution $u$ to this quadratic minimization problem on $V$, if it exists, must satisfy the
necessary condition 
$$
\mathbb{E}((R^\ast_\omega u-y,R^\ast_\omega w)_Y)=\mathbb{E}((P_{\omega}u-R_\omega y,w)_V)=0 \qquad \forall\; w\in V.
$$
This condition corresponds to the linear operator equation
\be\label{PRSchwarz}
P_\rho u=\mathbb{E}(R_\omega y), 
\qquad P_\rho:=\mathbb{E}(P_\omega)=\mathbb{E}(R_\omega R^\ast_\omega),
\ee
in $V$. Note that in general we cannot expect that $f_u=f_\mu$, since the closure $W$
of $H$ in $L^2_\rho(\Omega,Y)$ does not necessarily coincide with $L^2_\rho(\Omega,Y)$. 
For our convergence analysis in $V$, we require that
\be\label{AssR4}
\mathbb{E}(R_\omega y)\in \mathrm{ran}(P_\rho).
\ee
In other words, we assume that (\ref{Min2}) has a unique solution $u\in V$ for which (\ref{PRSchwarz}) holds. For general~$y$ with $\mathbb{E}(\|y\|_Y^2)<\infty$, only $\mathbb{E}(R_\omega y)\in \mathrm{ran}(P^{1/2}_\rho)$ can be established. It can also be proven that the functional $J(v)$ in (\ref{Min2})
does not have a global minimum if (\ref{AssR4}) is violated (the infimum of $J(v)$ is then only approached if $v\to \infty$ in $V$ in a certain way).
It will turn out that the condition (\ref{AssR4}) is necessary for the
convergence in $V$ of the online method (\ref{OnlineV})-(\ref{OnlineA}),
see Subsection \ref{sec42a}.

The operator $P_\rho:\,V\to V$ in (\ref{PRSchwarz}), which plays the role of a covariance operator, is bounded and symmetric positive definite.
Indeed, by definition we have
\bea
(P_\rho v,w)_V&:=&\int_{\Omega\times Y} (P_\omega v,w)_V\,d\mu(\omega,y)=\int_{\Omega} (P_\omega v,w)_V\,d\rho(\omega)\\
&=&\int_{\Omega} (R^\ast_\omega v,R^\ast_\omega w)_Y\,d\rho(\omega)=
(f_v,f_w)_{L^2_\rho(\Omega,Y)}=(v,P_\rho w)_V,\quad v,w\in V,
\eea
and 
$$
(P_\rho v,v)=\|f_v\|_{L^2_\rho(\Omega,Y)}^2\ge 0,\qquad v\in V.
$$ 
Even though it may happen that $P_\rho$ is not injective for certain $\rho$, we will from now on assume that $\mathrm{ker}(P_\rho) =\{0\}$ (otherwise, replace $V$ by its subspace $\{v\in V: \; \|f_v\|_{L^2_\rho(\Omega,Y)}=0 \}^\perp$).
The boundedness of $P_\rho: \,V\to V$, together with the estimate
$$
\|P_\rho\|_V\le \Lambda,
$$
follows from (\ref{AssR2}). The spectrum of $P_\rho$ is thus contained in $[0,\Lambda]$. 

In the formulation of the results and proofs below, we will additionally assume that $P_\rho$ is compact, which is the case in all known applications. A sufficient condition for compactness is the trace class property for $P_\rho$, which holds in particular if the operators $R_\omega$, $\omega\in \Omega$,  have uniformly bounded finite rank. In the scalar-valued case $Y=\mathbb{R}$, the trace class property holds if the associated  kernel function $k:\,\Omega\times \Omega \to \mathbb{R}$ satisfies
$$
\int_\Omega k(\omega,\omega)\,d\rho(\omega) <\infty,
$$  
which is automatically satisfied if the boundedness condition (\ref{AssR2a}) holds.
The compactness assumption enables us to define the scale of smoothness spaces $V_{P_\rho}^s$, $s\in \mathbb{R}$, generated by $P_\rho$ by simply using the complete orthonormal system (CONS) $\Psi:=\{\psi_k\}$ of eigenvectors of $P_\rho$ and associated eigenvalues $\lambda_1\ge\lambda_2\ge \ldots >0$ with
limit $0$ in $V$ as follows: $V_{P_\rho}^s$ is the completion  of $\mathrm{span}(\Psi)$
with respect to the norm
$$
\|\sum_k c_k \psi_k\|_{V_{P_\rho}^s} =\left( \sum_{k} \lambda_k^{-s}c_k^2 \right)^{1/2},
$$
which is well defined on $\mathrm{span}(\Psi)$ for any $s$. These spaces will appear in the investigation below.
For a noncompact $P_\rho$, the norm of the spaces $V_{P_\rho}^s$ can be defined using functional calculus, i.e., by replacing series representations using eigenfunction expansions by integrals with respect to the underlying
spectral family of the operator $P_\rho$. For example, this is worked out in \cite{EHN2000} in connection with iterative regularization methods of inverse problems (for $s>0$, the space  $V_{P_\rho}^s$ equals $\mathscr{X}_{s/2}$ defined in \cite[(3.29)]{EHN2000} if we replace $T^\ast T$ by $P_\rho$).

For a given prior RKHS $H$ induced by the feature map $\bfR$ with associated feature space $V$ and for given samples $(\omega_m,y_m)$, $m=0,\ldots,N-1$, with finite $N$,  the standard regularization of the ill-posed problem (\ref{Min1}) is to find the minimizer
$u_N\in V$ of the minimization problem
\be\label{Min3}
J_N(v):=\frac1N \sum_{m=0}^{N-1} \|f_v(\omega_m)- y_m\|_Y^2 + \kappa_N \|v\|_V^2\longmapsto \min
\ee
on $V$, where $\kappa_N>0$ is a suitable regularization parameter.
Using the representer theorem for Mercer kernels \cite{MiPo2005,CVT2006}, this problem leads to a  linear system with a typically dense and ill-conditioned $N\times N$ matrix. There is a huge body of literature, especially in the scalar-valued case $Y=\mathbb{R}$, devoted to setting up, analyzing and solving these problems for fixed $N$. Alternatively, one can restrict the minimization in (\ref{Min2}) to 
bounded, closed subsets in $V$ which, under our assumptions on $P_\rho$, are compact subsets of $L^2_\rho(\Omega,Y)$,
see \cite{CuZh2007}.

Here, we focus on the online learning algorithm (\ref{OnlineV})-(\ref{OnlineA}) for finding approximations to the minimizer $u\in V$ of (\ref{Min2}) and are interested in the analysis of their asymptotic performance.
For this, we define the noise term 
$$
\epsilon_\omega:=y-f_u(\omega)=y-R_{\omega}^\ast u,\qquad \omega\in \Omega,
$$
which is a $\mu$-distributed $Y$-valued random variable on $\Omega\times Y$ (to keep the notation short, the dependence of $\epsilon_\omega$ on $y$ is not explicitly shown).
By (\ref{AssR4}) we have $\mathbb{E}(\epsilon_\omega|\omega)=f_\mu(\omega)-f_u(\omega)$ for any $\omega\in \Omega$. Also, the noise variance
\be\label{OnlineVar}
\sigma^2_H:=\mathbb{E}(\|\epsilon_\omega\|_Y^2)=\mathbb{E}(\|y-f_\mu\|_Y^2)+\|f_\mu -f_u\|_{L^2(\Omega,Y)}^2
\ee
 with respect to $f_u\in H$ is finite, since $\mathbb{E}(\|y\|_Y^2)<\infty$ 
 was assumed in the first place. The value of $\sigma_H$ depends on both the average size of the noise $y-f_\mu(\omega)$ on $\Omega$ measured in the squared $Y$ norm and the $L_\rho^2(\Omega,Y)$ distance of $f_\mu$ from $W$.

The online algorithm (\ref{OnlineV})-(\ref{OnlineA}) is a particular instance of a randomized Schwarz iteration method associated with $\bfR$. Its noiseless version, where $y_m=R_{\omega_m}^\ast u$, was studied in \cite{GrOs2018} under the assumption that $u\in V^s_{P_\rho}$, $0\le s\le 1$, where $\alpha_m$ was as in (\ref{OnlineA}) but $\mu_m$ was determined by a steepest descent rule. Our goal in this paper is to derive convergence results for the expected squared error $\mathbb{E}(\|u-u^{(m)}\|^2_V)=\mathbb{E}(\|f_u-f^{(m)}\|^2_H)$, $m=1,2,\ldots$.
As expected, such estimates again require additional smoothness assumptions on $u$ in the form $u\in V_{P_\rho}^s$ with $0<s\le 1$. However, unlike the noiseless case \cite{GrOs2018}, these estimates also depend on the noise variance $\sigma_H^2$ in addition to the dependence on the initial error and the smoothness of $u$. The price of convergence
is a certain decay of the step-sizes $\mu_m\to 0$, as assumed in (\ref{OnlineA}), which is typical of 
stochastic approximation algorithms. 
Our main result is as follows:

\begin{theorem}\label{theo1} Let $Y,V$ be separable Hilbert spaces, $\Omega$ be a compact metric space, $\mu$ be a Borel probability measure on $\Omega\times Y$, and $\rho$ be the marginal Borel probability measure on $\Omega$ induced by $\mu$. Assume that
	$$
	\mathbb{E}(\|y\|_Y^2)=\int_{\Omega\times Y} \|y\|_Y^2 \,d\mu <\infty.
	$$
	For the feature map $\bfR=\{R_\omega\}_{\omega\in\Omega}$, we require uniform boundedness (\ref{AssR2})  and measurability. We also assume that the operator
	$P_\rho=\mathbb{E}(R_\omega R_\omega^\ast)$ is injective and compact.
	Finally, we assume (\ref{AssR4}) and that $u\in V_{P_\rho}^s$ for some $0<s\le 1$.
	
	Consider the online learning algorithm (\ref{OnlineV}), where $u^{(0)}\in V$ is arbitrary, the parameters $\alpha_m,\mu_m$ are given by (\ref{OnlineA}) with  $t=t_s:=(1+s)/(2+s)$ and $A=1/(2\Lambda)$, and 
	the random samples  $(\omega_m,y_m)$, $m=0,1,\ldots,N\le \infty$,
	are i.i.d. with respect to $\mu$.  Then the expected squared  error
	satisfies
	\be\label{OnlineEstF}
	\mathbb{E}(\|u-u^{(m)}\|^2_V)=\mathbb{E}(\|f_u-f^{(m)}\|_H^2)\le C^2(m+1)^{-s/(2+s)}, \qquad m=1,2,\ldots,N+1,
	\ee 
	where $f^{(m)}=f_{u^{(m)}}$, the noise variance $\sigma_H^2$ is defined in (\ref{OnlineVar}), and 
    $$
    C^2=2\|u-u^{(0)}\|^2_V+2\|u\|^2_V+8\Lambda^s\|u\|_{V_{P_\rho}^s}^2+\sigma_H^2/\Lambda.
    $$ 
\end{theorem}

In this generality, Theorem \ref{theo1} has not yet appeared in the literature, at least to our knowledge. Its proof is given in Section \ref{sec3}. For the parameter range $0<s\le 1$, the exponent   
$-s/(2+s)$ in the right-hand side of  (\ref{OnlineEstF}) is best possible under the general conditions stated in Theorem~\ref{theo1}, compare \cite{CaVi2007,LGZ2017}. 
Estimates of the form (\ref{OnlineEstF}) also  hold for arbitrary values
$1/2 < t <1$ and $0<A\le 1/(2\Lambda)$ in (\ref{OnlineA}), albeit with non-optimal exponents depending on $t$ and different constants $C$ varying with $t$ and $A$. Without the condition (\ref{AssR4}), which ensures the existence of the minimizer $u\in V$ in (\ref{Min2}), the online method (\ref{OnlineV})-(\ref{OnlineA}) diverges in expectation.
Note that convergence estimates with respect to the weaker $L^2_\rho(\Omega,Y)$ norm (hard learning scenario) cannot be obtained within our framework. We will comment on these issues in the concluding Section \ref{sec4}.

There is a huge amount of literature devoted to the convergence theory of various versions of
the algorithm (\ref{OnlineV})-(\ref{OnlineH}), especially for the 
scalar-valued case $Y=\mathbb{R}$. In particular, the algorithm is often considered in the so-called finite horizon case, where $N<\infty$ is fixed and the step-sizes $\mu_m$ are chosen in dependence on $N$ so that expectations such as $\mathbb{E}(\|u-u^{(N)}\|_{V}^2)$ or $\mathbb{E}(\|f_u-f_{u^{(N)}}\|_{L^2_\rho(\Omega,Y)}^2)$, respectively, are optimized for the final approximation $u^{(N)}$.  A brief discussion of known results is given in the next section.

\section{Examples and results related to Theorem \ref{theo1}}\label{sec2}
First, we provide some examples for the framework of this paper. We start with the scalar-valued case $Y=\mathbb{R}$
(or $Y=\mathbb{C}$), where suitable separable RKHS $H$ of functions $f:\, \Omega\to \mathbb{R}$ are often directly given by their associated bounded, symmetric, positive definite kernel function $k:\,\Omega\times\Omega\to \mathbb{R}$. Concrete examples of kernels, including Sobolev and Gaussian kernels, can be found in \cite{StCh2008,PaRa2016}. The canonical choice for the feature space $V$ is to identify $V=H$ and to define the feature map by
$$
R_\omega y=k(\omega,\cdot)y,\qquad R_\omega^\ast v=(k(\omega,\cdot),v)_H=v(\omega),\qquad y\in \mathbb{R},\quad v\in H,\quad \omega\in\Omega.
$$
As expected, the operator $P_\rho$ coincides with the integral operator
$$
(P_\rho v)(\omega) =\int_\Omega k(\omega,\theta) v(\theta)\,d\rho(\theta).
$$
We just note that the assumption (\ref{AssR4}) in Theorem \ref{theo1} is equivalent to
$$
\int_\Omega k(\omega,\theta) f_\mu(\theta)\,d\rho(\theta)\in \mathrm{ran}(P_\rho),
$$
which essentially means that $f_\mu$ must coincide with an element $v$ in the RKHS $H$ up to an additive perturbation from $\mathrm{ker}(\tilde{P}_\rho)$. Here $\tilde{P}_\rho$
denotes the extension of $P_\rho$ to all of $L^2_\rho(\Omega,\mathbb{R})$. In the literature, 
authors usually assume $f_\mu\in H=V_{P_\rho}^0$.
The choices of feature space and map are by no means unique. For example, one can set $V=\ell^2(\mathbb{N})$, fix an arbitrary complete orthonormal system (CONS) $(\phi_i)_{i\in\mathbb{N}}$ in $H$ and define the feature map by
$$
R_\omega y = (y\phi_i(\omega))_{i\in \mathbb{N}},\qquad R_\omega^\ast c = \sum_{i\in\mathbb{N}} c_i \phi_i(\omega), \qquad y\in \mathbb{R},\quad c:=(c_i)_{i\in\mathbb{N}}\in\ell^2(\mathbb{N}),\quad \omega\in \Omega.
$$
We do not go into further detail. Another example with $Y=\mathbb{R}$, where the RKHS $H$ can be identified with $\ell^2(\mathbb{N})$ will be considered in Subsection \ref{sec43}.

The construction of operator kernels $K$ for vector-valued RKHS typically involves one or several scalar-valued kernels $k$.
Here, we present an example of a multiplicative (or separable) operator kernel
\be\label{MultK}
K(\omega,\theta) = k(\omega,\theta)T,
\ee
where $T:\, Y\to Y$ is a bounded, positive definite, selfadjoint operator, which is a modification of \cite[Example 3]{KDPCRA2015}. Let
$Y=H^1_0(G)$, where $G$ is a fixed domain in $\mathbb{R}^d$ and set for simplicity $\Omega=[0,1]$. The associated learning problem is about finding a regression function $f_\mu:\, [0,1]\to H_0^1(G)$ with values in an infinite-dimensional Sobolev space.\footnote{A potential application could be finding approximations to the solution map $\omega\to u_\omega$ for the diffusion problem $-\nabla(a_\omega\nabla u) = f$ with random diffusion coefficient $a_\omega$, fixed
right-hand side $f$ and solution $u=u_\omega\in H^1_0(G)$ from samples $(\omega_i,u_i\approx u_{\omega_i})$.  Note that $Y$ could be any separable Hilbert space in the considerations below.}
For the feature space, consider the choice $V=L^2(\Omega)\otimes Y$ (this space can be identified with the Bochner space $L^2(\Omega,Y)$) and introduce the feature map by
$$
R_\omega u = (\chi_{[0,\omega)}-\omega)\otimes u, \qquad \omega\in \Omega,\quad u\in Y,
$$
where $\chi_E$ denotes the indicator function of a set $E\subset \Omega$.
Then, on elementary tensors, the adjoints $R_\omega^\ast$ are given by
$$
R_\omega^\ast (f\otimes v)=(\int_0^\omega f(t)\,dt -\omega\int_0^1 f(t)\,dt )v,\qquad f\in L^2(\Omega), \quad v\in Y.
$$
It takes some basic calculations to check that the associated operator kernel is of the form
$$
K(\omega,\theta)=k(\omega,\theta)I,
$$
where $I$ is the identity on $Y$ and 
$$
k(\omega,\theta)=\left\{\ba{ll} (1-\omega)\theta,&\;\theta\le \omega,\\
(1-\theta)\omega,&\; \theta\ge \omega, \ea\right.
$$
is the kernel for the scalar-valued RKHS $H_0^1(\Omega)$. Moreover, the resulting RKHS $H$ can be identified with $H^1_0(\Omega)\otimes Y$. Roughly speaking, functions in $H$ are of the form
$$
f(\omega)=\sum_{i\in\mathbb{N}} c_i(\omega)\phi_i \quad \mbox{with } c_i\in H^1_0(\Omega),
$$
while for functions in $L^2(\Omega,Y)$ such representations require only $c_i\in L_2(\Omega)$.
Here, $(\phi_i)_{i\in\mathbb{N}}$ is a CONS in $Y$.
In other words, convergence in $H$ implies a certain control of the smoothness of the coefficients $c_i(\omega)$ as functions on $\Omega$.

Given the large number of publications on convergence rates for learning algorithms,
we will present only a selection of results focusing on the RKHS setting and online algorithms similar to (\ref{OnlineV})-(\ref{OnlineH}). The results we cite are often stated and proved for the scalar-valued case $Y=\mathbb{R}$, although some authors claim that their methods extend to the case of an arbitrary separable Hilbert space $Y$ with minor modifications. One of the first papers on the vector-valued case is \cite{CaVi2007}, where the authors provide upper bounds in probability for the 
$L^2_\rho(\Omega,Y)$ norm of  $f_u-f_{u_N}$,
if $N\to \infty$ and $\kappa_N\to 0$, where $u_N$ is the solution of (\ref{Min3}). 
These bounds depend in a specific way on the smoothness of $u\in V_{P_\rho}^s$, $0\le s\le 1$, on compactness assumptions for the feature map $R_\omega$ and on the spectral properties of $P_\rho$. Note that 
the error measured in the $L^2_\rho(\Omega,Y)$ norm is with respect to $f_{u_N}$
and not with respect to approximations such as $f_{u^{(m)}}$, $m\le N$, which are produced by a special algorithm comparable to (\ref{OnlineH}). The results in \cite{CaVi2007} have recently  been  extended in \cite{FiSt2020,LMMG2024} to RKHS with multiplicative kernels (\ref{MultK}) to cover $V_{P_\rho}^t$ error bounds in probability under the assumption $f_\mu\in
V_{P_\rho}^s$, $-1\le t < s \le 1$. This covers $H$ ($t=0$)
and $L_\rho^2(\Omega,Y)$ ($t=-1$) convergence, including the hard learning scenario $f_\mu\not\in H$. The bounds require Bernstein-type inequalities for the noise level which hold in particular for uniformly bounded noise.

In \cite{TaYa2014}, the authors provide estimates in probability for an 
algorithm similar to (\ref{OnlineV})-(\ref{OnlineH}) in the scalar-valued case $Y=\mathbb{R}$. They cover both, convergence in $L^2_\rho(\Omega,\mathbb{R})$ and $H$ norms. There, the main additional assumption needed for the application of certain results from martingale theory is that, for some constant $M_\rho<\infty$, the random variable $y$ satisfies
$$
|y|\le M_\rho
$$
a.e. on the support of $\rho$.  If $u^{(0)}=0$ (as assumed in \cite{TaYa2014}), then
this assumption implies bounds for $\|u-u^{(0)}\|_V=\|u\|_V$
and $\sigma$ with constants that depend on
$M_\rho$. Up to the specification of constants and using the notation of this paper, the convergence result for the $H$ norm stated in  \cite[Theorem B]{TaYa2014} is as follows: Consider the online algorithm (\ref{OnlineV}) with starting value $u^{(0)}=0$ and parameters
$$
\alpha_m=\frac{m+m_0-1}{m+m_0},\qquad \alpha_m\mu_m=\frac{A}{(m+m_0)^{(s+1)/(s+2)}},\qquad m=0,1,\ldots,
$$
for some (sufficiently large) $m_0$ and suitable $A$. Then, if $u\in V_{P_\rho}^s$ for some
$0<s\le 2$, we have
$$
\mathbb{P}\left(\|u-u^{(m)}\|_V^2\le \frac{C}{(m+m_0)^{s/(s+2)}}\right)\ge 1-\delta,\qquad 0<\delta<1,\quad m=0,1,\ldots,
$$
for some constant 
$C=C(M_\rho,\|u\|_{V_{P_\rho}^s}, m_0, s, \Lambda, \log(2/\delta))<\infty$.
Here $V=H$  is an RKHS of functions $u:\,\Omega\to \mathbb{R}$ generated by some continuous scalar-valued Mercer kernel $k:\,\Omega\times \Omega
\to \mathbb{R}$ and  $\Lambda=\max_{\omega\in\Omega} k(\omega,\omega)$. 
So, for $0<s\le 1$, we get the same rate as in our Theorem \ref{theo1} which is, however, concerned with the expectation of the squared RKHS error in the more general vector-valued case. What our rather elementary method does not provide is a result for the case $1< s\le 2$ and for
$L^2_\rho(\Omega,Y)$ convergence. For the latter situation, \cite[Theorem C]{TaYa2014} gives the better estimate
$$
\mathbb{P}\left(\|u-u^{(m)}\|_{L^2_\rho(\Omega,\mathbb{R})}^2\le \frac{\bar{C}}{(m+m_0)^{(s+1)/(s+2)}}\right)\ge 1-\delta,\quad 0<\delta<1,\quad m=0,1,\ldots,
$$
under the same assumptions, but with a different constant 
$$
\bar{C}=\bar{C}(M_\rho,\|u\|_{V_{P_\rho}^s},m_0,s,\Lambda,\log(2/\delta))<\infty.
$$
This is almost matching the lower estimates for kernel learning derived in \cite{CaVi2007} for classes of instances where the spectrum of $P_\rho$ has a 
prescribed decay of the form $\lambda_k\asymp k^{-b}$ for some $b>1$, see also \cite{LGZ2017}.
Recall that the integral operator $P_\rho$ is trace class for scalar-valued Mercer kernels, while in our Theorem \ref{theo1} no stronger decay of eigenvalues is assumed.
The above cited bounds in probability automatically imply similar bounds for the expectation of squared errors.

Estimates in expectation close to our result have also been obtained for slightly different settings. For example, in \cite{YiPo2008} both, the so-called \emph{regularized} ($\alpha_m<1$) and the \emph{unregularized} online algorithm
($\alpha_m=1$) were analyzed in the scalar-valued case $Y=\mathbb{R}$ under assumptions similar to ours regarding $L^2_\rho(\Omega,\mathbb{R})$ and $V=H$ convergence. We quote only the result for convergence in the RKHS $V=H$. It concerns the so-called \emph{finite horizon} case of the unregularized online algorithm (\ref{OnlineV}) with $\alpha_m=1$, where one fixes $N<\infty$, chooses a constant step-size $\mu_m=\mu_N$,
$m=0,\ldots,N-1$, which depends on $N$, stops the iteration at $m=N$, and asks
for a good estimate of the expectation of $\mathbb{E}(\|u-u^{(N)}\|^2_V)$ for the final iterate only.
Up to the specification of constants, Theorem 6 in \cite{YiPo2008} states that, under the condition $u\in V^s_{P_\rho}$, $s>0$, one can obtain the bound
$$
\mathbb{E}(\|u-u^{(N)}\|^2_V)=\mathrm{O}(N^{-s/(s+2)}),\qquad N\to \infty,
$$
if one sets $\mu_N=cN^{-(s+1)/(s+2)}$ with a properly adjusted value of $c$. Note that $s>0$ is arbitrary with the exponent approaching $-1$ if the smoothness parameter $s$ tends to $ \infty$, while our result provides no improvement for $s>1$.  The drawback of the finite horizon case is that the estimate only concerns a fixed iterate $u^{(N)}$ with an $N$ to be decided beforehand. In a sense, this can be seen as building an approximation of the solution $u_N$ of (\ref{Min3}) with $\kappa_N=\mu_N$ from a single pass over the $N$ i.i.d. samples $(\omega_m,y_m)$, $m=0,\ldots,N-1$.

In recent years, attention has shifted to obtaining refined rates when $P_\rho$ possesses faster eigenvalue decay, usually expressed by the property that $P_\rho^\beta$ is trace class for some $\beta < 1$ or by the slightly weaker assumption 
\be\label{TrBetaWeak}
\lambda_k=\mathrm{O}(k^{-1/\beta}), \qquad k\to \infty,
\ee 
on the eigenvalues of the operator $P_\rho$.
Bounds that incorporate knowledge of $\beta<1$ are sometimes called capacity dependent, so our bound in Theorem~\ref{theo1} as well as the cited results from \cite{TaYa2014,YiPo2008} are capacity independent (in contrast, \cite{CaVi2007,FiSt2020,LMMG2024} all deal with capacity dependent estimates). Capacity dependent convergence rates for the expected squared error for the online algorithm
(\ref{OnlineV})-(\ref{OnlineH}) have been obtained, among others, in \cite{DiBa2016,DFB2017,GuSh2019,GLZ2022}, again in the scalar-valued case $Y=\mathbb{R}$ ($V=H$) and with various parameter settings, including unregularized and finite horizon versions. In \cite{DiBa2016}, rates for
$\mathbb{E}(\|u-\bar{u}^{(m)}\|_{L^2_\rho(\Omega,\mathbb{R})}^2)$ have been established,
where 
$$ 
\bar{u}^{(m)}=\frac1{m+1}\sum_{k=0}^m u^{(k)}, \qquad m=0,1,\ldots,
$$ 
is the sequence of averages associated with the sequence $u^{(m)}$, $m=0,1,\ldots$, obtained by the unregularized iteration (\ref{OnlineV}) with $\alpha_m=1$ and $u^{(0)}=0$.
That averaging has a similar effect as
regularization with $\alpha_m=(m+1)/(m+2)$ in (\ref{OnlineV}) considered in Theorem \ref{theo1} can be guessed if
one observes that 
$$
\bar{u}^{(m+1)} =\frac{m+1}{m+2}\bar{u}^{(m)} + \frac1{m+2}{u}^{(m+1)},
$$
where $u^{(m+1)}=u^{(m)}+\mu_m(y_{m}-R_{\omega_m}u^{(m)})$, and compares with 
the regularized iteration (\ref{OnlineV}) with $\alpha_m$ given in (\ref{OnlineA}). 
To illustrate the influence of $\beta$, we formulate the following bound, which is a 
consequence of \cite[Corollary 3]{DiBa2016}: Under an additional technical assumption on the noise term $\epsilon_\omega$, if the condition (\ref{TrBetaWeak}) holds for some $0<\beta<1$ and
$u\in V_{P_\rho}^s$, $s>-1$, then for suitable choices for the learning rates $\mu_m$, we have
$$
\mathbb{E}(\|u-\bar{u}^{(m)}\|_{L^2_\rho(\Omega,\mathbb{R})}^2) = 
\left\{\ba{ll} \mathrm{O}((m+1)^{-(s+1)}),\quad& -1<s< -\beta, \\ \mathrm{O}((m+1)^{-(s+1)/(s+1+\beta)}),\quad& -\beta<s< 1-\beta, \\
\mathrm{O}((m+1)^{-(1-\beta/2)}),\quad& 1-\beta <s.  \ea\right.
$$
Thus, stronger eigenvalue decay implies stronger asymptotic error decay in the $L^2_\rho(\Omega,\mathbb{R})$ norm. In \cite[Section 6]{DFB2017},
similar rates are obtained in the finite horizon setting for both, the above averaged iterates 
$\bar{u}^{(N)}$ and for $u^{(N)}$ produced by a two-step extension 
of the one-step iteration (\ref{OnlineV}).

In addition to $L^2_\rho(\Omega,\mathbb{R})$ convergence results, \cite{GuSh2019} also provides a capacity dependent convergence estimate in the RKHS norm for the unregularized algorithm (\ref{OnlineV})-(\ref{OnlineH}) with parameters $\alpha_m=1$ and
$\mu_m=c(m+1)^{-1/2}$. Under the boundedness assumption $|y|\le M_\rho$, Theorem 2 in \cite{GuSh2019} implies 
that 
$$
\mathbb{E}(\|u-{u}^{(m)}\|^2_V)=\mathrm{O}((m+1)^{-\min(s,1-\beta)/2}\log^2(m+1)),
\qquad m=1,2,\ldots,
$$
if $u\in V^s_{P_\rho}$ for some $s>0$, $P_\rho^\beta$ is trace class for some 
$0<\beta<1$, and $c$ is properly adjusted. 

Finally, the scalar-valued kernel regression problem with $Y=\mathbb{R}$ and prior RKHS $V=H$ can also be cast as linear regression
problem in $V=H$. This was done in \cite{DFB2017,GLZ2022}. More abstractly,
given a $\mu$-distributed random variable $(\xi_\omega,y)\in V\times \mathbb{R}$ on $\Omega\times \mathbb{R}$, the task is to find approximations to the minimizer $u\in V$ of the problem
$$ 
\mathbb{E}(|(\xi_\omega,v)_V-y|^2)\longmapsto \min,\qquad v\in V,
$$ 
from i.i.d. samples $(\xi_{\omega_i},y_i)$. If $k$ is the scalar-valued kernel of the RKHS $V=H$ 
then the canonical choice is $\xi_\omega=k(\omega,\cdot)$. In \cite{GLZ2022}, weak convergence in $V$ is studied for the
iteration
$$
u^{(m+1)} =u^{(m)} + \mu_m (y_m-(\xi_{\omega_m},u^{(m)}))\xi_{\omega_m},\qquad m=0,1,\ldots,
$$
by deriving estimates for quantities such as $\mathbb{E}((v,u-u^{(m)})_V^2)$ and $\mathbb{E}((\xi_{\omega'},u-u^{(m)})_V^2)$ under some assumptions on the learning rates $\mu_m$, a more restrictive noise model, and the normalization  $\|\xi_\omega\|_V=1$. 
This iteration is nothing but the unregularized iteration (\ref{OnlineV}) with $\alpha_m=1$, since $(\xi_{\omega_m},u^{(m)})_V=u^{(m)}(\omega_m)$ in the scalar-valued case. Note that 
the assumption $\|\xi_\omega\|_V=1$ means $k(\omega,\omega)=1$. 
Moreover, in this case
$$
\mathbb{E}((\xi_{\omega'},u-u^{(m)})_V^2)=\mathbb{E}(\|u-u^{(m)}\|_{L^2_\rho(\Omega,\mathbb{R})}),
$$
since the expectation on the left, in addition to the i.i.d. samples $(\xi_{\omega_i},y_i)$, $i=0,\ldots,m-1$, is also taken with respect to the independently $\rho$-distributed  random variable $\xi_{\omega'}$. This implies learning rates in the $L^2_\rho(\Omega,\mathbb{R})$ norm. The  estimates for $\mathbb{E}((\xi_{\omega'},u-u^{(m)})_V^2)$ given in \cite{GLZ2022} concern both, the finite horizon and the online setting and again  depend on the parameters 
$s\ge 0$ (smoothness of $u$) and $0<\beta\le 1$ (capacity assumption on $P_\rho$). Convergence in the RKHS norm is studied in the finite horizon case, see \cite[Corollary 2.8]{GLZ2022}. For the estimates of $\mathbb{E}((v,u-u^{(m)})_V^2)$, the smoothness $s'\ge 0$ of the fixed element $v\in V^{s'}_{P_\rho}$ is traded against the smoothness $s\ge 0$ of $u\in V^{s}_{P_\rho}$, see \cite{GLZ2022} for details.

\section{Proof of Theorem \ref{theo1}}\label{sec3}
In this subsection we will use the notation and assumptions outlined above, with the only change that the scalar product in $V$ is simply denoted by $(\cdot,\cdot)$ and the associated norm $\|\cdot\|_V$ is accordingly denoted by $\|\cdot\|$. We also set $e^{(m)} := u-u^{(m)}$.
We will prove an estimate of the form
\be\label{OnlineEst0}
\mathbb{E}(\|e^{(m)}\|^2)=\mathrm{O}((m+1)^{-s/(2+s)}),\qquad m\to \infty,
\ee
under the assumption $u\in V_{P_\rho}^s$, $0<s\le 1$, if the parameters $A$ and $t$
in (\ref{OnlineA}) are chosen properly. The precise statement and the dependence
of the constant in (\ref{OnlineEst0}) on the initial error, the noise variance and the smoothness assumption
are given in the formulation of Theorem \ref{theo1}. 

From (\ref{OnlineV}) and $y_m=R_{\omega_m}^\ast u + \epsilon_{\omega_m}$ we derive the error representation
$$
e^{(m+1)}=\underbrace{\alpha_m(e^{(m)}-\mu_m P_{\omega_m} e^{(m)}) +\bar{\alpha}_m u}_{\bar{e}^{(m+1)}:=} - \alpha_m\mu_m R_{\omega_m}\epsilon_{\omega_m},
$$
where $\bar{\alpha}_m:=1-\alpha_m=(m+2)^{-1}$, compare also (\ref{OnlineA}). The first term $\bar{e}^{(m+1)}$ corresponds to the noiseless case considered in \cite{GrOs2018}, while the remainder term is the noise contribution.
Thus,
\be\label{OnlineErr1}
\|e^{(m+1)}\|^2=\|\bar{e}^{(m+1)}\|^2-2\alpha_m\mu_m (R_{\omega_m}\epsilon_{\omega_m},\bar{e}^{(m+1)}) + \alpha_m^2\mu_m^2 \|R_{\omega_m}\epsilon_{\omega_m}\|^2.
\ee

We now estimate the conditional expectation with respect to the given $u^{(m)}$, separately for the three terms in (\ref{OnlineErr1}). Here and in the following we denote this conditional expectation by $\mathbb{E}'$. 
For the third term, by (\ref{AssR2}) and the definition (\ref{OnlineVar}) of the noise variance $\sigma_H^2$, we have
\be\label{OnlineEst1a}
\mathbb{E}'(\|R_{\omega_m}\epsilon_{\omega_m}\|^2)\le \Lambda\mathbb{E}(\|\epsilon_{\omega}\|_Y^2)=\Lambda \sigma_H^2.
\ee
For the second term, we need
$$
\mathbb{E}((R_{\omega}\epsilon_{\omega},w))=\mathbb{E}((y-R_{\omega}^\ast u,R_{\omega}^\ast w)_Y)=0\qquad \forall \;w\in V.
$$
This follows directly from the fact that $u\in V$ is the minimizer of the problem (\ref{Min2}). Thus, by setting $w=\alpha_m e^{(m)}+\bar{\alpha}_m u$, we get
\bea
&&\mathbb{E}'(-2\alpha_m\mu_m(R_{\omega_m}\epsilon_{\omega_m},\bar{e}^{(m+1)}))\\ && \qquad\quad =2\alpha_m\mu_m(\alpha_m\mu_m\mathbb{E}'((R_{\omega_m}\epsilon_{\omega_m},P_{\omega_m} e^{(m)}))-\mathbb{E}'( (R_{\omega_m}\epsilon_{\omega_m},w)))\\
&&\qquad\quad = 2\alpha_m^2\mu_m^2\mathbb{E}'((R_{\omega_m}\epsilon_{\omega_m},P_{\omega_m} e^{(m)}))\\
&&\qquad\quad \le \alpha_m^2\mu_m^2(\mathbb{E}'(\|R_{\omega_m}\epsilon_{\omega_m}\|^2)+\mathbb{E}'(\|P_{\omega_m} e^{(m)}\|^2)).
\eea
Here, the first term is estimated by (\ref{OnlineEst1a}). For the second term, we substitute
the upper bound
\be\label{OnlineEst1b}
\mathbb{E}'(\|P_{\omega_m} e^{(m)}\|^2)\le \Lambda\mathbb{E}'((P_{\omega_m} e^{(m)},e^{(m)})) = \Lambda(P_\rho e^{(m)},e^{(m)}),
\ee
which follows from (\ref{AssR2}) and the definition of $P_\rho$. Together this gives
\be\label{OnlineEst1c}
\mathbb{E}'(-2\alpha_m\mu_m (R_{\omega_m}\epsilon_{\omega_m},\bar{e}^{(m+1)}))\le \Lambda \alpha_m^2\mu_m^2 (\sigma_H^2+(P_\rho e^{(m)},e^{(m)}))
\ee
for the second term in (\ref{OnlineErr1}).

To estimate the first term $\mathbb{E}'(\|\bar{e}^{(m+1)}\|^2)$, we modify the arguments from \cite{GrOs2018}, where the case $\epsilon_m=0$ was treated. We use the error decomposition
\bea
\|\bar{e}^{(m+1)}\|^2&=& \bar{\alpha}_m^2\|u\|^2 +2\alpha_m \bar{\alpha}_m(u,e^{(m)}-\mu_m P_{\omega_m}e^{(m)}\\
&&\quad\quad\quad\quad\quad\quad\quad\quad +\, \alpha_m^2(\|e^{(m)}\|^2-2\mu_m(e^{(m)},P_{\omega_m}e^{(m)}))
+\mu_m^2\|P_{\omega_m}e^{(m)}\|^2).
\eea
After taking conditional expectations, we arrive with the definition of $P_\rho$
and (\ref{OnlineEst1b}) at
\bea
\mathbb{E}'(\|\bar{e}^{(m+1)}\|^2)&=& \bar{\alpha}_m^2\|u\|^2 +2\alpha_m \bar{\alpha}_m(u,e^{(m)}-\mu_m P_\rho e^{(m)})\\
&&\quad +\, \alpha_m^2(\|e^{(m)}\|^2-2\mu_m(e^{(m)},P_\rho e^{(m)})
+\mu_m^2\mathbb{E}'(\|P_{\omega_m}e^{(m)}\|^2))\\
&\le & \bar{\alpha}_m^2\|u\|^2 +2\alpha_m \bar{\alpha}_m(u,e^{(m)}-\mu_m P_\rho e^{(m)})\\
&&\quad +\, \alpha_m^2(\|e^{(m)}\|^2-\mu_m(2-\Lambda\mu_m)(e^{(m)},P_\rho e^{(m)})).
\eea
Next, to estimate the term
$(u,e^{(m)}-\mu_m P_\rho e^{(m)})$,
we take an arbitrary 
$h=P^{1/2}_\rho v\in V_{P_\rho}^1$, where $v\in V=V_{P_\rho}^0$ and
$\|h\|_{V_{P_\rho}^1}=\|v\|$. This gives us
\bea
&& 2\alpha_m \bar{\alpha}_m(u,e^{(m)}-\mu_m P_\rho e^{(m)})\\
&&\quad = 2\alpha_m \bar{\alpha}_m((u-h,(I-\mu_m P_\rho) e^{(m)})+(h,(I-\mu_m P_\rho) e^{(m)}))\\
&&\quad \le 2\alpha_m \bar{\alpha}_m\|u-h\|\|(I-\mu_m P_\rho) e^{(m)}\| +2  (\bar{\alpha}_m\mu_m^{-1/2}(I-\mu_m P_\rho)v,\alpha_m\mu_m^{1/2}e^{(m)})\\
&&\quad \le 2\alpha_m \bar{\alpha}_m\|u-h\| \|e^{(m)}\| + \bar{\alpha}_m^2\mu_m^{-1}\|(I-\mu_m P_\rho)v\|^2 + \alpha_m^2\mu_m\|P_\rho^{1/2}e^{(m)}\|^2\\
&&\quad \le 2\alpha_m \bar{\alpha}_m\|u-h\| \|e^{(m)}\| +\bar{\alpha}_m^2\mu_m^{-1}\|h\|_{V_{P_\rho}^1}^2 + \alpha_m^2\mu_m (P_\rho e^{(m)},e^{(m)}).
\eea
Here we have silently used that $\|(I-\mu_m P_\rho)e^{(m)}\|\le \|e^{(m)}\|$
and similarly
$$
\|(I-\mu_m P_\rho)v\|\le \|v\|=\|h\|_{V_{P_\rho}^1},
$$ 
which holds since $0<\mu_m\le A\le (2\Lambda)^{-1}$ according to (\ref{OnlineA}) and the restriction on $A$.
Substitution into the previous inequality yields
\bea
\mathbb{E}'(\|\bar{e}^{(m+1)}\|^2)&\le& 
\bar{\alpha}_m^2(\|u\|^2 +\mu_m^{-1}\|h\|_{V_{P_\rho}^1}^2)+2\alpha_m \bar{\alpha}_m \|u-h\| \|e^{(m)}\|\\
&&\quad +\, \alpha_m^2(\|e^{(m)}\|^2-\mu_m(1-\Lambda\mu_m)(e^{(m)},P_\rho e^{(m)})).
\eea

Now, combining this estimate  for the conditional expectation of the first term in (\ref{OnlineErr1}) with the bounds (\ref{OnlineEst1a}) and (\ref{OnlineEst1c}) for the third and second terms, respectively, we get
\be\label{OnlineEst1}
\mathbb{E}'(\|\bar{e}^{(m+1)}\|^2)\le \alpha_m^2(\|e^{(m)}\|^2+2\Lambda\sigma_H^2\mu_m^2)\qquad\qquad\qquad\qquad\quad
\ee
$$
\qquad\qquad\qquad\qquad +\, 2\alpha_m \bar{\alpha}_m \|u-h\| \|e^{(m)}\| + \bar{\alpha}_m^2(\|u\|^2 +\mu_m^{-1}\|h\|_{V_{P_\rho}^1}^2).
$$
Here the term involving $(e^{(m)},P_\rho e^{(m)})\ge 0$ has been omitted, since
its resulting forefactor $-\mu_m(1-2\Lambda\mu_m)$ is non-positive due to the restriction on
$A$ in (\ref{OnlineA}).

For given 
$$
u=\sum_k c_k\psi_k \in V_{P_\rho}^s,\qquad 0<s\le 1,
$$
in (\ref{OnlineEst1}) we choose
$$
h=\sum_{k:\,\lambda_k (m+1)^b\ge B} c_k \psi_k
$$
with some fixed constants $b,B>0$ specified below. This gives
$$
\|h\|_{V^1_{P_\rho}}^2=\sum_{k:\,\lambda_k (m+1)^b\ge B} \lambda_k^{-(1-s)} (\lambda_k^{-s}c_k^2)\le B^{-(1-s)}(m+1)^{(1-s)b}\|u\|_{V_{P_\rho}^s}^2
$$
and
$$
\|u-h\|^2=\sum_{k:\,\lambda_k (m+1)^b < B} \lambda_k^{s} (\lambda_k^{-s}c_k^2)\le B^{s}(m+1)^{-bs}\|u\|_{V_{P_\rho}^s}^2.
$$
After substitution into (\ref{OnlineEst1}), we get
\be\label{OnlineEst2}
\mathbb{E}'(\|\bar{e}^{(m+1)}\|^2)\le \alpha_m^2(\|e^{(m)}\|^2+2\Lambda\sigma_H^2\mu_m^2) + 2\alpha_m \bar{\alpha}_m B^{s/2}(m+1)^{-bs/2}\|u\|_{V_{P_\rho}^s} \|e^{(m)}\| 
\ee
$$
\qquad\qquad +\, \bar{\alpha}_m^2(\|u\|^2 +\mu_m^{-1}B^{-(1-s)}(m+1)^{(1-s)b}\|u\|_{V_{P_\rho}^s}^2).
$$
Obviously, if $s=1$, we can set $h=u$ which greatly simplifies the 
considerations below and leads to a more accurate final estimate, see Section \ref{sec41}.

Next, we switch to full expectations in (\ref{OnlineEst2}) by using the independence assumption for the sampling process and by taking into account that 
$$
\varepsilon_m:=\mathbb{E}(\|e^{(m)}\|^2)^{1/2}\ge \mathbb{E}(\|e^{(m)}\|).
$$
Together with (\ref{OnlineA}) and $\alpha_m=(m+1)\bar{\alpha}_m$,
this gives
\bea
\varepsilon_{m+1}^2&\le& \alpha_m^2(\varepsilon_m^2+2A^2\Lambda\sigma_H^2(m+1)^{-2t}) + 2\alpha_m \bar{\alpha}_m B^{s/2}(m+1)^{-bs/2}\|u\|_{V_{P_\rho}^s}\varepsilon_m  \\
&&\qquad + \,\bar{\alpha}_m^2(\|u\|^2 +A^{-1}B^{-(1-s)}(m+1)^{(1-s)b+t}\|u\|_{V_{P_\rho}^s}^2)\\
&\le& \alpha_m^2\varepsilon_m^2+\bar{\alpha}_m^2(2A^2\Lambda\sigma_H^2(m+1)^{2-2t} + 2B^{s/2}(m+1)^{-bs/2+1}\|u\|_{V_{P_\rho}^s}\varepsilon_m \|e^{(m)}\| \\
&&\qquad  +\, \|u\|^2 +A^{-1}B^{-(1-s)}(m+1)^{(1-s)b+t}\|u\|_{V_{P_\rho}^s}^2).
\eea
In a final step, we assume for a moment that
\be\label{OnlineEst3}
\varepsilon_k\le C(k+1)^{-r},\qquad k=0,\ldots,m,
\ee
holds for some constants $C,r>0$. Next, we set 
$$
a:=\max(2-2t,-bs/2+1-r,(1-s)b+t)
$$
and
$$
D:=2A^2\Lambda\sigma_H^2+2CB^{s/2}\|u\|_{V_{P_\rho}^s}+\|u\|^2+A^{-1}B^{-(1-s)}\|u\|_{V_{P_\rho}^s}^2.
$$
Since $1/2 < t  <1$ is assumed in (\ref{OnlineA}), we have $a>0$.
Then, for $k=0,1,\ldots,m$, the estimate for $\varepsilon_{k+1}$ simplifies to
$$
\varepsilon_{k+1}^2\le \alpha_k^2 \varepsilon_k^2 +D\bar{\alpha}_k^2(k+1)^{a}
$$
or, since  $\alpha_k^2\bar{\alpha}_{k-1}^2=\bar{\alpha}_k^2$, to
$$
d_{k+1}:=\bar{\alpha}_k^{-2}\varepsilon_{k+1}^2\le \alpha_k^2\bar{\alpha}_k^{-2} \varepsilon_k^2 +D(k+1)^{a} = d_k +D(k+1)^{a}.
$$
By recursion we get
$$
d_{m+1} \le d_0+D\sum_{k=0}^m (k+1)^a =\varepsilon_0^2+D\sum_{k=0}^m (k+1)^a,
$$
and finally 
$$
\varepsilon_{m+1}^2\le (m+2)^{-2}(\|e^{(0)}\|^2+D(m+2)^{a+1})<(\|e^{(0)}\|^2+D)(m+2)^{a-1},
$$
since we have $a > 0$ and
\be\label{Sum}
\sum_{k=0}^m (k+1)^a \le \int_1^{m+2} x^a\,dx < (m+2)^{a+1}.
\ee
So (\ref{OnlineEst3}) holds by induction for all $m$ if we ensure that
\be\label{RecCond}
1-a\ge 	2r,\qquad \|e^{(0)}\|^2+D\le C^2.
\ee

To complete the proof of Theorem \ref{theo1}, it remains to maximize $r$ for given $0< s \le 1$. For this purpose, it is intuitively clear to require 
$$
a=1-2r=2-2t=-bs/2+1-r=(1-s)b+t.
$$
This system of equations has the unique solution
$$
t=\frac{1+s}{2+s},\quad b=\frac{1}{2+s}, \quad 2r=\frac{s}{2+s}, \quad a=\frac{2}{2+s}.
$$
Furthermore, the appropriate value for $C$ in (\ref{OnlineEst3}) must satisfy 
$$
 \|e^{(0)}\|^2+\|u\|^2+2A^2\Lambda\sigma_H^2+2CB^{s/2}\|u\|_{V_{P_\rho}^s}+A^{-1}B^{-(1-s)}\|u\|_{V_{P_\rho}^s}^2\le C^2.
$$
With such choices for $t$ and $C$, the condition (\ref{RecCond}) is guaranteed, and (\ref{OnlineEst3}) gives the desired bound 
$$
\varepsilon_m^2\le C^2(m+1)^{-s/(s+2)},\qquad m=1,2,\ldots,N-1.
$$
By choosing concrete values for $0<A\le (2\Lambda)^{-1}$ and $B>0$, the constant $C^2$ can be made more explicit. For example, substituting the upper bound
$$
2CB^{s/2}\|u\|_{V_{P_\rho}^s}\le \frac{C^2}2 +2B^s\|u\|_{V_{P_\rho}^s}^2
$$
and rearranging the terms shows that
$$
C^2=2\left(\|e^{(0)}\|^2+\|u\|^2+B^s(2+(AB)^{-1})\|u\|_{V_{P_\rho}^s}^2+2A^2\Lambda\sigma_H^2\right)
$$
is appropriate. In particular, setting for simplicity  $A$ to its maximal value $A=(2\Lambda)^{-1}$ and
taking $B=\Lambda$ gives a more explicit dependence of $C^2$ on
the assumptions on $\|e^{(0)}\|^2$, the noise variance $\sigma_H^2$, and the smoothness of $u$, namely
$$ 
C^2=2\|e^{(0)}\|^2+2\|u\|^2+8\Lambda^s\|u\|_{V_{P_\rho}^s}+\sigma_H^2/\Lambda.
$$ 
This is the constant shown in the formulation of Theorem \ref{theo1}. Obviously, varying $A$ and $B$ changes the trade-off between initial error,  noise variance, and smoothness assumptions in
the convergence estimate (\ref{OnlineEst3}). Note also that $B$ is not part of the algorithm and can be adjusted to any value. Finally, the analysis of (\ref{RecCond}) shows that the bound (\ref{OnlineEstF}) holds with some constant $C$ and the exponent $s/(s+2)$ replaced by $\min(2t-1,(1-t)s)$ for arbitrary $1/2<t<1$ and $0<A\le (2\Lambda)^{-1}$ in (\ref{OnlineA}),
see Subsection \ref{sec41} for details in the case $s=1$.
This concludes the proof of Theorem \ref{theo1}.

\section{Further remarks}\label{sec4} 
\subsection{Comments on Theorem \ref{theo1}}\label{sec41}
In the special case $s=1$, the proof of Theorem \ref{theo2} is simplified as follows: In (\ref{OnlineEst1}) we can set $h=u$, and (\ref{OnlineEst2}) is therefore simplified to
$$ 
\mathbb{E}'(\|\bar{e}^{(m+1)}\|^2)\le \alpha_m^2(\|e^{(m)}\|^2+2\Lambda\sigma_H^2\mu_m^2) + \bar{\alpha}_m^2(\|u\|^2 +\mu_m^{-1}\|u\|_{V_{P_\rho}^1}^2).
$$ 
So with $\mu_m=A(m+1)^{-t}$ we directly get a recursion for $$
d_m:=\bar{\alpha}_{m-1}^{-2}\varepsilon_m^2=(m+1)^2\mathbb{E}(\|e^{(m)}\|^2)
$$
in the form  
$$
d_{m+1} \le d_m +(2A^2\Lambda\sigma_H^2(m+1)^{2-2t} +\|u\|^2+A^{-1}(m+1)^{t}\|u\|_{V_{P_\rho}^1}^2).
$$
Taking (\ref{Sum}) into account, we finally arrive for $1/2<t<1$ at
$$ 
\mathbb{E}(\|e^{(m)}\|^2)\le \frac{\|e^{(0)}\|^2}{(m+1)^2}+\frac{2A^2\Lambda\sigma_H^2}{(m+1)^{2t-1}}+
\frac{\|u\|^2}{m+1} +\frac{A^{-1}\|u\|_{V_{P_\rho}^1}^2}{(m+1)^{1-t}},
$$ 
$m=1,2,\ldots$. This estimate shows more clearly the guaranteed error decay with respect to the initial error $\|e^{(0)}\|^2$, the noise variance $\sigma_H^2$, and the norms $\|u\|^2$ and $\|u\|_{V_{P_\rho}^1}^2$ of the solution $u$ in dependence on $t$. The asymptotically
dominant term is of the form 
$\mathrm{O}((m+1)^{-\min(2t-1,1-t)})$ 
and is minimized when $t=2/3$. For this value of $t$ and with $A=(2\Lambda)^{-1}$ we get
\be\label{OnlineEstnew1}
\mathbb{E}(\|e^{(m)}\|^2)\le \frac{\|e^{(0)}\|^2}{(m+1)^2}+\frac{\|u\|^2}{m+1}+\frac{2\Lambda\|u\|_{V_{P_\rho}^1}^2+(2\Lambda)^{-1} \sigma_H^2}{(m+1)^{1/3}}.
\ee
Without further assumptions one cannot expect a better error decay rate, see Section \ref{sec2} and Subsection \ref{sec43}.

Another comment concerns the finite horizon setting, which is often treated instead of a true online method. Here one fixes a finite 
$N$, chooses a constant learning rate $\mu_m=\mu$ for $m=0,\ldots,N-1$ in dependence on $N$, and only asks for a best possible bound for $\mathbb{E}(\|e^{(N)}\|^2)$. Our approach easily provides 
results for this case as well. We demonstrate this only for $s=1$.
For fixed $\mu_m=\mu$, the error recursion for the quantities $d_m$ now takes the form
$$
d_{m+1} \le d_m +(2A^2\Lambda\sigma_H^2(m+1)^{2}\mu^2 +\|u\|^2+\mu^{-1}\|u\|_{V_{P_\rho}^1}^2),\quad m=0,\ldots,N-1,
$$
and gives
$$
\mathbb{E}(\|e^{(N)}\|^2)\le \frac{\|e^{(0)}\|^2}{(N+1)^2}+2\Lambda\sigma_H^2\mu^2(N+1) +
\frac{\|u\|^2+\mu^{-1}\|u\|_{V_{P_\rho}^1}^2}{N+1}.
$$
Setting $\mu=(2\Lambda)^{-1}(N+1)^{-2/3}$ results in a final estimate for the finite horizon case similar to 
(\ref{OnlineEstnew1}), but only for $m=N$.

There are obvious drawbacks of the whole setting in which Theorem \ref{theo1} is formulated.
First, the assumptions are at most qualitative: Since $\mu$, and thus $\rho$, is usually not at our disposal, we cannot verify the assumption $u\in V_{P_\rho}^s$, nor assess the value of $\sigma_H^2$.
Moreover, although the restriction to learning rates $\mu_m$ of the form (\ref{OnlineA}) may not cause problems in view of the results obtained, the choice of optimal values for $t$ and $A$ is by no means obvious. It would be desirable to have
a rule for the adaptive choice of $\mu_m$ that does not require knowledge of the values of $s$ and the size of the norms of $u$, but leads to the same quantitative error decay as guaranteed by Theorem \ref{theo1}.

\subsection{Difficulties with convergence in $L^2_\rho(\Omega,Y)$}\label{sec42}
Our result for the vector-valued case concerned convergence in
$V$, which is identical with the RKHS $H$ generated by $\bfR$.  What we did not succeed in is extending our methods to obtain better asymptotic convergence rates of $f_{u^{(m)}} \to f_u$ in the $L^2_\rho(\Omega,Y)$ norm. Under
 the assumption (\ref{AssR4}) about the existence of the minimizer $u$ in (\ref{Min2}), error estimates in the $L^2_\rho(\Omega,Y)$ norm require the study of $\mathbb{E}(\|P_\rho^{1/2} e^{(m)}\|^2)=\mathbb{E}((P_\rho e^{(m)},e^{(m)}))$ instead of $\mathbb{E}(\|e^{(m)}\|^2)$.
If, in analogy to (\ref{OnlineErr1}), one examines the error decomposition
\bea
\|P_\rho^{1/2} e^{(m+1)}\|^2\le \|P_\rho^{1/2} \bar{e}^{(m+1)}\|^2
-2\alpha_m\mu_m (P_\rho R_{\omega_m}\epsilon_{\omega_m},\bar{e}^{(m+1)})
+ \alpha_m^2\mu_m^2 \|P_\rho^{1/2}R_{\omega_m}\epsilon_{\omega_m}\|^2,
\eea
then difficulties mostly arise from the first term in the right-hand side. In fact, we have
\bea
\|P_\rho^{1/2} \bar{e}^{(m+1)}\|^2&=& \bar{\alpha}_m^2\|P_\rho^{1/2}u\|^2 +2\alpha_m \bar{\alpha}_m(P_\rho u,e^{(m)}-\mu_m P_{\omega_m}e^{(m)})\\
&+&\alpha_m^2(\|P_\rho^{1/2}e^{(m)}\|^2-2\mu_m(P_\rho e^{(m)},P_{\omega_m}e^{(m)})
+\mu_m^2\|P_\rho^{1/2}P_{\omega_m}e^{(m)}\|^2).
\eea
After taking conditional expectations $\mathbb{E}'(\|P_\rho^{1/2} \bar{e}^{(m+1)}\|^2)$, we get a negative term
$$
-2\alpha_m^2 \mu_m \|P_\rho e^{(m)}\|^2
$$
on the right-hand side, which must compensate for positive contributions from terms such as
$$
\mathbb{E}'(\|P_\rho^{1/2} P_{\omega_m}e^{(m)}\|^2).
$$
Since in general $P_\rho$ does not commute with the operators $P_\omega$, it is not clear how to relate these quantities 
without additional assumptions.

\subsection{A divergence result}\label{sec42a} If the crucial condition (\ref{AssR4}) in Theorem~\ref{theo1} does not hold, i.e., if 
$$ 
g:=\mathbb{E}(R_\omega y)=\mathbb{E}(R_\omega f_\mu(\omega))\not\in \mathrm{ran}(P_\rho),
$$ 
then the sequence $u^{(m)}$ obtained from the online algorithm
(\ref{OnlineV})-(\ref{OnlineA}) diverges in expectation to $\infty$ in $V$ for any choice of the parameters $1/2 <t<1$ and $0<A\le (2\Lambda)^{-1}$. This negative result is equivalent to proving
\be\label{OnlineDiv}
\|\mathbb{E}(u^{(m)})\|^2 \to \infty,\qquad m\to \infty,
\ee
and shows that (\ref{AssR4}) is essential in Theorem \ref{theo1}. As before, norm and scalar product in $V$ are denoted by $\|\cdot\|$ and $(\cdot,\cdot)$, respectively. 

To establish (\ref{OnlineDiv}), we expand $g$ and $u^{(0)}$ with respect to the CONS $\Psi$ and derive an inhomogeneous linear recursion for the coefficients of the expected error trajectory $U^{(m)}:=\mathbb{E}(u^{(m)})$. To do this, set
$$
u^{(0)}=\sum_{i\in \mathbb{N}} c_i\psi_i,\qquad g=\sum_{i\in \mathbb{N}} g_i\psi_i,
$$
where $c_i=(u^{(0)},\psi_i)$, $g_i=(g,\psi_i)$, and denote $c_i^{(m)}:=(U^{(m)},\psi_i)$, $i\in \mathbb{N}$. Obviously, $u^{(0)}=U^{(0)}$ and thus $c_i=c_i^{(0)}$. From (\ref{OnlineV}) we have
$$
U^{(m+1)}=\alpha_m(U^{(m)}+\mu_m(g-P_\rho U^{(m)}))=\alpha_m(I-\mu_mP_\rho)U^{(m)}+\alpha_m\mu_m g,\quad m\ge 0,
$$
which, by iteration using the properties of the regularization parameters $\alpha_m=(m+1)/(m+2)$, immediately yields
$$
U^{(m)}=\frac1{(m+1)}\left(\prod_{l=0}^{m-1}(I-\mu_l P_\rho)u^{(0)} + \sum_{k=1}^m k\mu_{k-1}\prod_{l=k}^{m-1}(I-\mu_l P_\rho)g\right),
$$
$m=1,2,\ldots$. By linearity of the expectation operator and the fact that $\Psi$ consists of the eigenvectors of $P_\rho$, we get
\be\label{ciIter}
c_i^{(m)}=\frac1{(m+1)}\left(\underbrace{\prod_{l=0}^{m-1}(1-\mu_l \lambda_i)}_{\epsilon_m(\lambda_i)}c_i + \underbrace{(\sum_{k=1}^m k\mu_{k-1}\prod_{l=k}^{m-1}(1-\mu_l \lambda_i))}_{d_m(\lambda_i):=}g_i\right),
\ee
$m=1,2,\ldots$, separately for each $i\in \mathbb{N}$.
Under the restrictions on the parameters $\mu_m$ from
(\ref{OnlineA}), we have $1/2\le 1-\mu_l \lambda_i \le 1$. This shows that we have
$0<\epsilon_m(\lambda_i)\le 1$ for the factor in front of $c_i$, which implies that the contribution
of the initial guess $u^{(0)}$ can be neglected if $m\to\infty$. 

Next, we focus on lower bounds for the factor $d_m(\lambda_i)$ in front of $g_i$
in (\ref{ciIter}). For our purposes it is sufficient to show that, for some
$m_0\ge 1$ depending on $t$,
\be\label{dmlambda}
d_m(\lambda)\ge C_0 (m+1)\lambda^{-1}, \qquad C_1 (m+1)^{t-1}\le \lambda \le \Lambda,\qquad m\ge m_0,
\ee
with constants $C_0$, $C_1$ independent of $\lambda$ and $m$. In fact, with (\ref{dmlambda}) at hand, we have
\bea
\|U^{(m)}\|^2&=& (m+1)^{-2}\sum_{i\in\mathbb{N}} (\epsilon_m(\lambda_i)c_i +d_m(\lambda_i)g_i)^2\\
&\ge&(m+1)^{-2}\left(\frac12\sum_{i\in\mathbb{N}} d_m(\lambda_i)^2g_i^2     -\sum_{i\in\mathbb{N}} \epsilon_m(\lambda_i)^2c_i^2\right)\\
&\ge&\frac{C_0^2}2  \left(\sum_{\lambda_i\ge C_1(m+1)^{t-1}} \lambda_i^{-2}g_i^2 \right)- \frac{\|u^{(0)}\|^2}{(m+1)^2},
\eea
where the elementary inequality $(a+b)^2\ge b^2/2-a^2$, $a,b\in\mathbb{R}$, was used in the first step.
But if $g\not\in \mathrm{ran}(P_\rho)$, then $\sum_{i\in\mathbb{N}} \lambda_i^{-2}g_i^2=\infty$ and the above lower bound also tends to infinity. This proves (\ref{OnlineDiv}).

It remains to show (\ref{dmlambda}). For this, set
$$
\Pi_k^{m-1}(\lambda):=\prod_{l=k}^{m-1} (1- \mu_k\lambda),\quad k=0,\ldots,m-1,\qquad\Pi_m^{m-1}(\lambda):=1.
$$
Since $1/2\le 1-\mu_m\lambda\le 1$, we have $\log(1-\mu_m\lambda)\ge -\log (2)\mu_m\lambda$
for all $m\ge 0$. This gives
$$
\Pi_k^{m-1}(\lambda)\ge 2^{-\lambda\sum_{l=k}^{m-1} \mu_l}, \qquad k=1,\ldots,m-1.
$$
But for $0<t<1$ we have
\bea
\sum_{l=k}^{m-1}\mu_l &=&A\sum_{l=k+1}^m l^{-t}\le \frac1{2\Lambda}\int_{k+1/2}^{m+1/2} x^{-t}\,dx\\
&=&\frac{(m+\frac12)^{1-t}-(k+\frac12)^{1-t}}{2\Lambda(1-t)}\le C (m+1)^{1-t}-(k+1)^{1-t}),
\quad k=1,\ldots,m-1,
\eea
where $C$ depends on $t$ and $\Lambda$. So,
$$
\Pi_k^{m-1}(\lambda)\ge 2^{-C\lambda((m+1)^{1-t}-(k+1)^{1-t})}, \qquad k=1,\ldots,m,
$$
and, consequently,
\be\label{dmlambda1}
d_m(\lambda)\ge \frac12 \sum_{k\in I_{m}(\lambda)} k \mu_{k-1}, \qquad I_m(\lambda):=\{k\le m:\;C\lambda((m+1)^{1-t}-(k+1)^{1-t})\le 1\}.
\ee
Obviously, the cardinality of $I_m(\lambda)$ is equal to $|I_m(\lambda)|=m-k_0$, where
$k_0$ is the largest $k\ge 0$ not in $I_m(\lambda)$, i.e.,
$$
(k_0+1)^{1-t} < (m+1)^{1-t}-(C\lambda)^{-1} \le  (k_0+2)^{1-t}.
$$ 
Now, if 
$$
(2/C)(m+1)^{t-1}\le\lambda \le\Lambda,
$$
then $(k_0+2)^{1-t}\ge (m+1)^{1-t}/2$, which implies that, for sufficiently large $m\ge m_0$, there exists such a $k_0$ that satisfies $k_0+1\ge C_2(m+1)$ with some constant $C_2>0$, where $m_0$ and $C_2$ depend on $t$. Furthermore, by definition of $k_0$, we have
$$
\frac1{C\lambda} < (m+1)^{1-t}-(k_0+1)^{1-t} \le (1-t)\frac{m-k_0}{(k_0+1)^t}.
$$
When we substitute this, $k\ge k_0+1\ge C_2(m+1)$, and $\mu_{k-1}=Ak^{-t}$ into (\ref{dmlambda1}), we get
\bea
d_m(\lambda)&\ge& \frac{A}2 (k_0+1)^{1-t}(m-k_0)\ge \frac{A(k_0+1)}{2(1-t)C\lambda}\\
&\ge& \frac{AC_2}{2(1-t)C}\frac{m+1}{\lambda}, \qquad (2/C)(m+1)^{t-1}\le\lambda \le\Lambda,\quad m\ge m_0.
\eea
This proves (\ref{dmlambda}) if we set 
$$
C_0=\frac{AC_2}{2(1-t)C},\qquad C_1=\frac2{C},
$$
and finishes the argument for (\ref{OnlineDiv}).

\subsection{A special case}\label{sec43}
Now consider the special "learning" problem of recovering an unknown element
$u\in V$ from noisy measurements of its coefficients with respect to a CONS $\Psi=\{\psi_i\}_{i\in \mathbb{N}}$ in $V$ by the online method considered in this paper. To do this, we assume that
we are given $\mu$-distributed random samples $(i_m,y_m)$, where $i_m\in \mathbb{N}$ and
$$ 
y_m=(u,\psi_{i_m}) + \epsilon_m,\qquad m=0,1,\ldots
$$ 
are the noisy samples of the coefficients $(u,\psi_i)$. Starting from $u^{(0)}=0$,
we want to approximate $u$ by the iterates $u^{(m)}$ obtained from the online algorithm
\be\label{OnlineCONSAlgo}
u^{(m+1)}=\alpha_m u^{(m)} + \alpha_m\mu_m(y_m-(u^{(m)},\psi_{i_m}))\psi_{i_m},\qquad m=0,1,\ldots,
\ee
where the coefficients $\alpha_m$ and $\mu_m$ are given by (\ref{OnlineA}) with $\Lambda=1$. This is a special instance of (\ref{OnlineV}) if we set $\Omega=\mathbb{N}$, $Y=\mathbb{R}$ and
define $R_i:\,\mathbb{R}\to V$ and  $R^\ast_i:\,V\to\mathbb{R}$ by
$R_iy=y\psi_i$ and $R_i^\ast v=(v,\psi_i)$, respectively. The associated RKHS $H$ can be identified with $\ell^2(\mathbb{N})$. To simplify things further, let $i_m$ be i.i.d.
samples from $\mathbb{N}$ with respect to a discrete probability measure $\rho$ on $\mathbb{N}$,
and let $\epsilon_m$ be i.i.d. random noise with zero mean and finite variance $\sigma^2<\infty$ that is independent of $i_m$. 
The corresponding operator $P_\rho$ is given by
$$
P_\rho v = \sum_{i\in\mathbb{N}} \rho_i (v,\psi_i)\psi_i,
$$
its eigenvalues $\lambda_i=\rho_i$ are given by $\rho$, and it is trace class
(w.l.o.g., we assume $\rho_1\ge\rho_2\ge \ldots$). The spaces $V_{P_\rho}^s$, $-\infty<s<\infty$, can now be identified as sets of formal orthogonal series
$$
V_{P_\rho}^s:=\left\{u \sim \sum_{i\in\mathbb{N}} c_i\psi_i\; :\quad \|u\|_{V_{P_\rho}^s}^2=\sum_{i\in\mathbb{N}} \rho_i^{-s}c_i^2\right\}.
$$
Obviously, $V_{P_\rho}^s \subset V=V_{P_\rho}^0$ for $s>0$. Since functions $f:\,\mathbb{N}\to \mathbb{R}$ can be identified with formal series with respect to $\Psi$ by 
$$
u  \sim \sum_{i\in\mathbb{N}} c_i\psi_i \quad \leftrightarrow \quad f_u:\;f_u(i)=c_i,
$$
we have $\|f_u\|_{L^2_\rho(\mathbb{N},\mathbb{R})} = \|u\|_{V_{P_\rho}^{-1}}$ and we can
silently identify $L^2_\rho(\mathbb{N},\mathbb{R})$ with $V_{P_\rho}^{-1}$. Under the assumptions made, the underlying minimization problem  (\ref{Min2}) on $V$ is
$$
\mathbb{E}(|f_v-y|^2)=\|f_v-f_u\|_{L^2_\rho(\mathbb{N},\mathbb{R})}^2+\sigma^2\quad \longmapsto \quad \min,
$$
and has $u$ as its unique solution. This example also shows that sometimes it is natural to consider convergence in $V$ rather than convergence in $L_\rho^2(\Omega,Y)$.

The simplicity of this example allows a comprehensive convergence theory with respect to the scale of $V_{P_\rho}^s$ spaces. We state the following results without proof.
\begin{theorem}\label{theo2} Let $\,-1\le \bar{s}\le 0\le s$, and $\bar{s}<s\le \bar{s}+2$. Then, for the sampling process described above, the online algorithm  (\ref{OnlineCONSAlgo}) converges for $u\in V_{P_\rho}^s$ in the $V_{P_\rho}^{\bar{s}}$ norm with the bound
\be\label{OnlineCONS}
\mathbb{E}(\|e^{(m)}\|^2_{V_{P_\rho}^{\bar{s}}})\le C(m+1)^{-\min((s-\bar{s})/(s+2),2/(\bar{s}+4))}
( A^{\bar{s}-s} \|u\|^2_{V_{P_\rho}^s}   +  A^{2+\bar{s}} \sigma^2), 
\ee
$m=1,2,\ldots$, if the parameters $t$ and $A$ in (\ref{OnlineA}) satisfy  
$$
t=t_{s,\bar{s}}:=\max((s+1)/(s+2),(\bar{s}+3)/(\bar{s}+4)), \qquad 0<A\le 1/2. 
$$
\end{theorem}

Setting $\bar{s}=0$, one concludes from (\ref{OnlineCONS}) that the  
convergence estimate for the online algorithm (\ref{OnlineV}), which holds
by Theorem \ref{theo1} for $0 < s \le 1$ in the general case, is indeed matched.
For $\bar{s}=-1$, which corresponds to $L^2_\rho(\mathbb{N},\mathbb{R})$ convergence, the rate is better and in line with known lower bounds. 

The estimate (\ref{OnlineCONS}) for the online algorithm (\ref{OnlineCONSAlgo}) is best possible in the sense that, under the conditions of Theorem \ref{theo2}, the exponent in (\ref{OnlineCONS})  cannot be increased without additional assumptions on $\rho$. In particular, there is no further improvement for $s>\bar{s}+2$, i.e., the estimate actually saturates at $s=\bar{s}+2$. This can be seen from the following result.
\begin{theorem}\label{theo3} Let $-1\le \bar{s} \le 0 \le s$,  $\bar{s} < s$ and $\sigma>0$. For the online algorithm (\ref{OnlineCONSAlgo}) we have 
$$ 
\sup_\rho \sup_{u:\,\|u\|_{V_{P_\rho}^s}=1 } (m+1)^{\min((s-\bar{s})/(2+s),2/(\bar{s}+4))}\mathbb{E}(\|e^{(m)}\|^2_{V_{P_\rho}^{\bar{s}}})\ge c> 0,  
$$ 
$m=1,2,\ldots,$, where $c$ depends on $\bar{s}$, $s$, $\sigma$, and the parameters $t$ and $A$ in (\ref{OnlineA}), but is independent of $m$. 
\end{theorem}

The proofs of these statements are elementary but rather tedious and will be given elsewhere.
Let us just note that the simplicity of this example allows us to reduce the considerations
to explicit linear recursions for expectations associated with the decomposition coefficients $c_i^{(m)} :=(e^{(m)},\psi_i)$ of the errors $e^{(m)}=u-u^{(m)}$ with respect to $\Psi$ for each $i\in \mathbb{N}$ separately. This is because
\be\label{OnlineCONS2}
\mathbb{E}(\|e^{(m)}\|^2_{V_{P_\rho}^{\bar{s}}})=\sum_i \rho_i^{-\bar{s}}\mathbb{E}((c_i^{(m)})^2),\qquad \|u\|^2_{V_{P_\rho}^s}=\|e^{(0)}\|^2_{V_{P_\rho}^s}=\sum_i \rho_i^{-s}c_i^2
\ee
and
\bea
c_i^{(m+1)} &=&\bar{\alpha}_m c_i + \alpha_m c_i^{(m)}+\alpha_m \left\{\ba{ll}\mu_m (y_{i_m}-(u^{(m)},\psi_{i_m})),&\; i_m=i\\ 0,&\;i_m\neq i \ea \right.\\
&=& \bar{\alpha}_m c_i + \alpha_m(c_i^{(m)}-\delta_{i,i_m}\mu_m (c_i^{(m)}+\epsilon_m))
\eea
for $m=0,1,\ldots$, where $\delta_{i,i_m}=1$ with probability $\rho_i$, and $\delta_{i,i_m}=0$ with probability $1-\rho_i$. So if we denote $\varepsilon_{m,i}:=\mathbb{E}((c_i^{(m)})^2)$ and $\bar{\varepsilon}_{m,i}:=\mathbb{E}(c_i^{(m)})$ and use the independence assumption, we get a system of linear recursions
\bea
\varepsilon_{m+1,i}&=&\alpha_m^2(1-\rho_i\mu_m(2-\mu_m))\varepsilon_{m,i} + 2\alpha_m\bar{\alpha}_m(1-\rho_i\mu_m) c_i \bar{\varepsilon}_{m,i} 
 + \,\bar{\alpha}_m^2 c_i^2 + \rho_i\alpha_m^2\mu_m^2 \sigma^2,\\
\bar{\varepsilon}_{m+1,i}&=&\alpha_m(1-\rho_i\mu_m)\bar{\varepsilon}_{m,i} + \bar{\alpha}_m c_i,
\eea
$m=0,1,\ldots$, with starting values $\varepsilon_{0,i}=c_i^2$ and $\bar{\varepsilon}_{0,i}=c_i$. In principle, this system can be solved explicitly. For example, we have
$$
\bar{\varepsilon}_{m,i}=\frac1{m+1}c_iS_m,\qquad S_m:=\sum_{k=0}^m \Pi_{k}^{m-1},
$$
where the notation
$$ 
\Pi_k^{m-1}:=(1-\frac{a}{m^t})\cdot \ldots \cdot (1-\frac{a}{(k+1)^t}), \quad 0\le k\le m-1, \quad \Pi_m^{m-1}:=1,
$$ 
is used with $a=A\rho_i$. Similarly, we get
$$
\varepsilon_{m,i} \le\frac1{(m+1)^2}\left(2c_i^2\sum_{k=0}^{m}\Pi_{k}^{m-1}S_k + \rho_i\sigma^2 \sum_{k=0}^{m-1} (k+1)^2\mu_k^2\Pi_{k+1}^{m-1}\right).
$$
A matching lower bound for $\varepsilon_{m,i}$ can be obtained by using a slightly different value of $a$ in the definition of 
the products $\Pi_k^{m-1}$. The remainder of the argument for Theorem \ref{theo2} first requires the substitution of tight upper
bounds for $\Pi_k^{m-1}$ and $S_k$ in dependence on $t$ and $a$ into the bounds for $\varepsilon_{m,i}$.
Next, after substituting the estimates for $\varepsilon_{m,i}$ into (\ref{OnlineCONS2}), the resulting
series has to be estimated separately for the index sets $I_1:=\{i:\,A\rho_i\le (m+1)^{t-1}\}$ and $I_2:=\mathbb{N}\backslash I_1$ followed by choosing the indicated optimal value of $t=t_{s,\bar{s}}$.
This leads to the bound (\ref{OnlineCONS}) in Theorem \ref{theo2}. For the proof of Theorem \ref{theo3}, lower bounds are needed for 
$\Pi_k^{m-1}$, $S_k$, and consequently $\varepsilon_m$, combined with choosing
suitable discrete probability distributions  $\rho$. Regarding lower bounds for
$\Pi_k^{m-1}$, see the considerations in  Subsection \ref{sec42a}.

\section*{Acknowledgement}
Michael Griebel and Peter Oswald were supported by the {\em Hausdorff Center for Mathematics} in Bonn, funded by the Deutsche Forschungsgemeinschaft (DFG, German Research
Foundation) under Germany's Excellence Strategy - EXC-2047/1 - 390685813 and the CRC 1060 {\em The Mathematics of Emergent Effects} of the Deutsche Forschungsgemeinschaft.

\bibliographystyle{amsplain}

\end{document}